%% file: main.tex
\documentclass[letterpaper, 10 pt, conference]{ieeeconf}  % Comment this line out if you need a4paper

\IEEEoverridecommandlockouts                              % This command is only needed if 
\usepackage{amsfonts}
\usepackage{amsmath}
\usepackage{algorithm}
\usepackage{algpseudocode}
\usepackage[dvipsnames]{xcolor}
\usepackage{setspace}
\usepackage{graphicx}
\usepackage{bbm}
\usepackage{url}
\usepackage{hyperref}
\usepackage{booktabs}
\usepackage{multirow, makecell}
\usepackage{lettrine}
\usepackage{makecell}
\usepackage{caption}
\usepackage{subcaption}
\usepackage{dblfloatfix}
\usepackage{tabularx}

\definecolor{fig_magenta}{RGB}{255,0,255}
\definecolor{fig_blue}{RGB}{0,133,188}
\definecolor{fig_red}{RGB}{193,39,45}
\definecolor{fig_orange}{RGB}{241,90,36}

\DeclareMathOperator*{\argmin}{arg\,min}

\renewcommand{\baselinestretch}{.99}
\newcommand{\actions}{\mathcal{A}}
\newcommand{\freeactions}{\actions_{free}}

\newcommand{\rgbspace}{\mathcal{O}_{RGB}}

\newcommand{\latentspace}{\mathcal{X}}
\newcommand{\graph}{\mathcal{G}}

\newcommand{\obs}{o}
\newcommand{\code}{x}
\newcommand{\datatraj}{\tau_o}
\newcommand{\actiontraj}{\tau_a}

\newcommand{\posmargin}{m_{+}}
\newcommand{\negmargin}{m_{-}}

\newcommand{\local}{h}
\newcommand{\conn}{f^{\dagger}}
\newcommand{\fd}{f}
\newcommand{\georeg}{p^{+}}
\newcommand{\norm}[1]{\left\lVert#1\right\rVert_2}
\newcommand{\localmetric}{d_{\local}}

\title{\LARGE \bf One-4-All: Neural Potential Fields for Embodied Navigation}

\author{Sacha Morin$^{*1}$, Miguel Saavedra-Ruiz$^{*1}$ and Liam Paull$^{1}$% <-this % stops a space
\thanks{$^{*}$Equal contribution. Author ordering determined by competitive duck-calling, where the winner was selected by a blind jury on their ability to recreate various duck calls.}%
\thanks{$^{1}$Sacha Morin, Miguel Saavedra-Ruiz and Liam Paull are with the Department of Computer Science and Operations Research, Université de Montréal,
        Montréal, QC, Canada, and with Mila - Quebec AI Institute, Montréal, QC, Canada.
        {\tt\small sacha.morin@mila.quebec}}%
}
\begin{document}

\maketitle
\thispagestyle{empty}
\pagestyle{empty}

%%%%%%%%%%%%%%%%%%%%%%%%%%%%%%%%%%%%%%%%%%%%%%%%%%%%%%%%%%%%%%%%%%%%%%%%%%%%%%%%
\begin{abstract}
A fundamental task in robotics is to navigate between two locations. In particular, real-world navigation can require long-horizon planning using high-dimensional RGB images, which poses a substantial challenge for end-to-end learning-based approaches. Current semi-parametric methods instead achieve long-horizon navigation by combining learned modules with a topological memory of the environment, often represented as a graph over previously collected images. However, using these graphs in practice requires tuning a number of pruning heuristics. These heuristics are necessary to avoid spurious edges, limit runtime memory usage and maintain reasonably fast graph queries in large environments. In this work, we present One-4-All (O4A), a method leveraging self-supervised and manifold learning to obtain a graph-free, end-to-end navigation pipeline in which the goal is specified as an image. Navigation is achieved by greedily minimizing a potential function defined continuously over image embeddings. Our system is trained offline on non-expert exploration sequences of RGB data and controls, and does not require any depth or pose measurements. We show that O4A can reach long-range goals in 8 simulated Gibson indoor environments and that resulting embeddings are topologically similar to ground truth maps, even if no pose is observed. We further demonstrate successful real-world navigation using a Jackal UGV platform.\footnote{Project page \url{https://montrealrobotics.ca/o4a/}.}

\end{abstract}

%%%%%%%%%%%%%%%%%%%%%%%%%%%%%%%%%%%%%%%%%%%%%%%%%%%%%%%%%%%%%%%%%%%%%%%%%%%%%%%%
\section{INTRODUCTION}

Navigation is a crucial component in any robotics stack that requires a robot to move from one location to another. This problem is characterized by a robot's ability to identify the most efficient and feasible path between a start pose and a goal pose in a given environment. The standard approach involves first piloting the robot within the environment to build a metric map, often using a range sensor, and then using this representation for planning \cite{thrun2005robotics}. However, the memory complexity of these methods scales poorly with the size of the environment, and they do not exploit semantic information nor visual cues \cite{shah2021rapid}. 

As an alternative, learning-based approaches, also dubbed \textit{experiential learning} \cite{Levine2023survey}, have gained momentum due to their ability to work directly with high-dimensional data (e.g., images) and reason about non-geometric concepts in a scene. Furthermore, these methods are more intuitive to use for non-expert users as they allow for goal positions to be specified using images of places or objects rather than coordinates in a metric space \cite{chaplot2020NTS}. However, end-to-end experiential learning generally learns a global controller that maps images directly to actions, failing to reason about long-horizon goals. Furthermore, they are well-known for being data inefficient \cite{dulan2021challenges, shah2022revind}.

To overcome the challenges of long-horizon navigation, \textit{topological} memory representations \cite{Kuipers1988classic} are used to divide the navigation problem into two parts. First, the memory representation is used to produce a globally coherent navigation plan, which is then followed waypoint by waypoint using a learned or classical local controller \cite{bansal2020combining}. Approaches that incorporate both memory and learning-based components are referred to as \textit{semi-parametric}, while approaches that rely solely on learning are known as \textit{fully-parametric}.  

While semi-parametric methods have proven to be effective for image-based navigation both indoors \cite{savinov2018sptm, wiyatno2022lifelong, chaplot2020NTS, kim2022topological} and outdoors \cite{shah2021ving, shah2022viking}, they still encounter memory issues. This is a result of the topological memory typically being encoded as a graph whose nodes represent visited states and edges represent traversability. As the environment size increases, more nodes and edges are required in the graph, increasing the memory requirements. 

\begin{figure}
\medskip
\centerline{
\subfloat[First trajectory]{\includegraphics[width=.48\columnwidth]{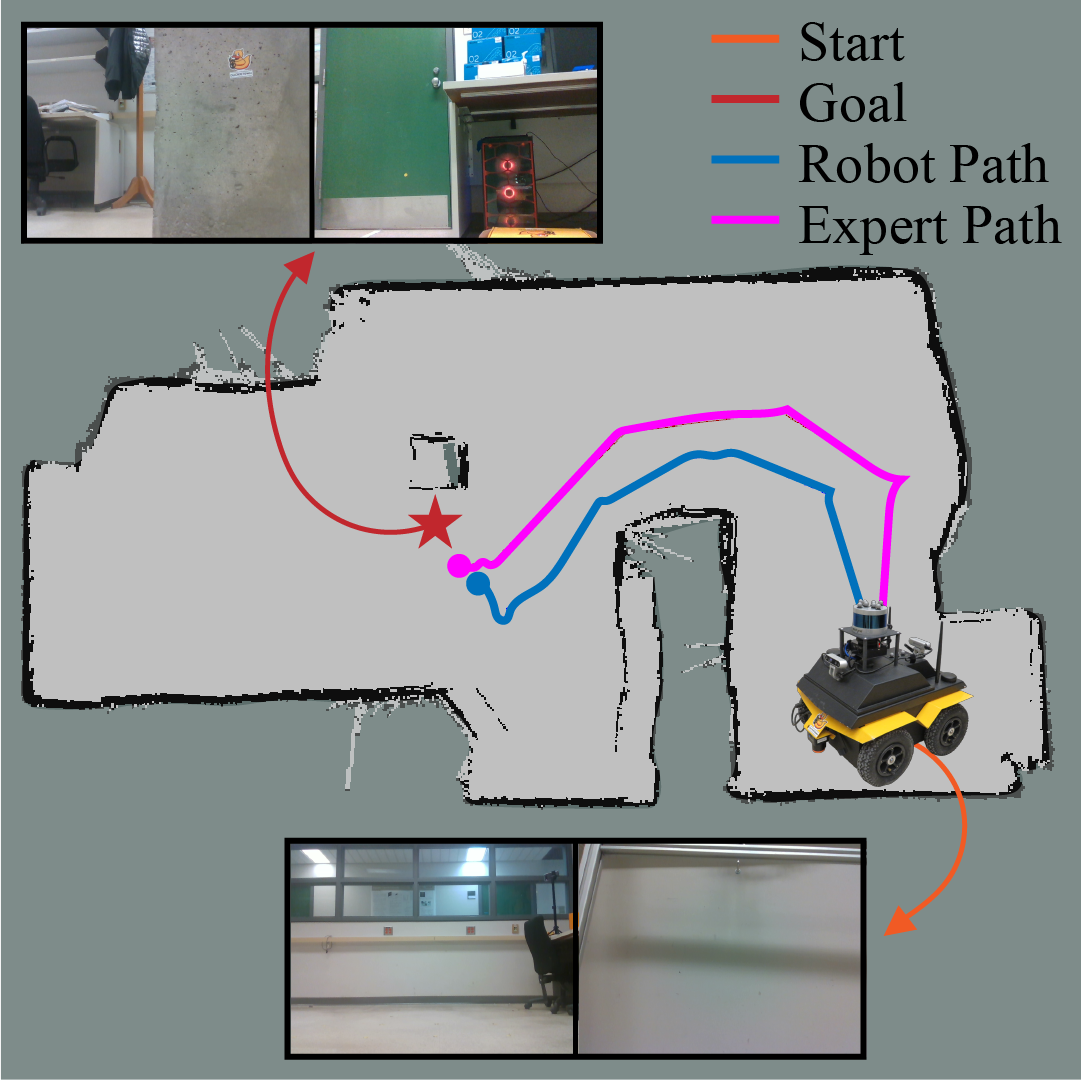}}
\hfil
\subfloat[Second trajectory]{\includegraphics[width=.48\columnwidth]{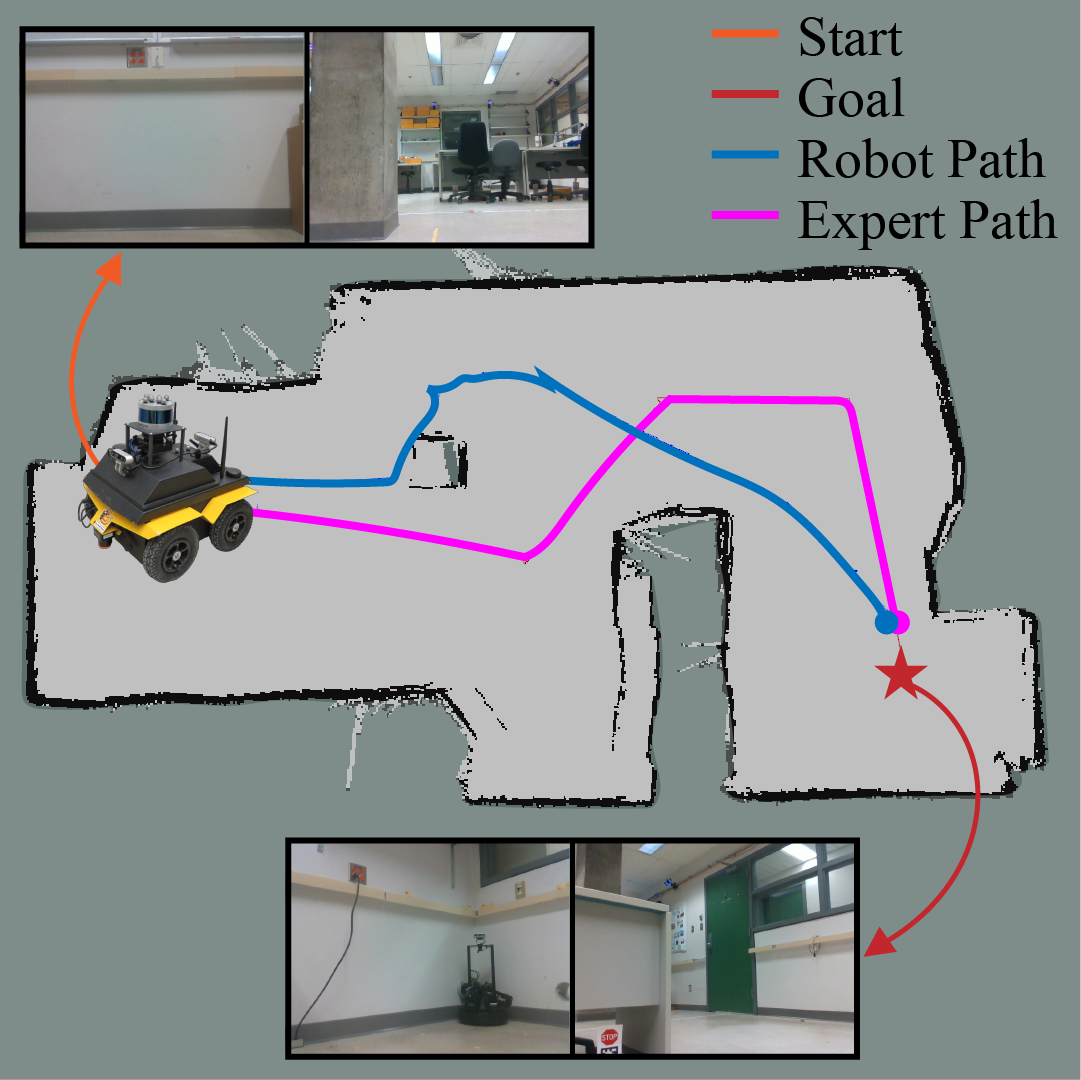}}
}
% \vspace{5.0mm}
\caption{Navigation paths produced by O4A ({Robot, \color{fig_blue} blue}) and by human teleoperation (Expert, {\color{fig_magenta} magenta}) in a 4.65m x 9.10m laboratory. When prompted with a goal image ({\color{fig_red} red star}), the robot uses its current front and back RGB observations ({\color{fig_orange} orange}) to navigate towards the goal by minimizing a neural potential function over image embeddings. The paths were captured using information from a ViCON system, which was not available to the agent.}
\label{fig:robot_trajs}
\end{figure}

In addition, spurious connections in the graph can impede navigation performance as they may represent non-feasible transitions in the physical world, leading to failure modes in the global planning stage. Although the literature offers partial solutions to these limitations by pruning the graph with hand-crafted heuristics \cite{wiyatno2022lifelong, shah2021ving}, these methods add complexity to the problem and generally require tuning for each environment.

To address these limitations, we propose \textbf{One-4-All} (O4A), an end-to-end fully-parametric method for image-goal navigation. O4A is trained offline using non-expert exploration sequences of RGB data and controls. We first rely on self-supervised learning to identify neighboring RGB observations. Armed with this notion of connectivity, we compute a graph to derive a manifold learning objective \cite{joshua2000isomap, yang2020plan2vec} for our planning module, which we dub the \textbf{geodesic regressor}. The geodesic regressor will learn to predict shortest path lengths between pairs of RGB images and in that sense, encodes the geometry of the environment and acts as our memory module \cite{gupta2017cognitive, georgakis2022learningsemantic, henriques2018mapnet, kuan2019memory}.  While we do compute a temporary graph during training, we discard it for navigation, and found that it does not require  the hand-crafted graph pruning heuristics of existing semi-parametric methods. Intuitively, we trade a potentially high number of nodes and edges in a graph for a fixed number of learnable parameters, thus mitigating the memory limitations of semi-parametric approaches. 
Inference is also improved: graph queries are replaced with efficient forward passes in a neural network.

For navigation, we draw inspiration from potential fields planning \cite{choset2005principles} and use the output of our geodesic regressor as an attractor in a potential function. This allows us to frame navigation as a minimization problem, with the global minimizer being the goal image. We show how this navigation approach enables the robot to perform long-horizon navigation and succeed even in geometrically complex environments.

To summarize, our main contributions are:
\begin{itemize}
    \item An offline self-supervised training procedure using non-expert exploration sequences of RGB data and controls, without any depth or pose measurements;
    \item A graph-free, end-to-end navigation pipeline that avoids tuning graph pruning heuristics;
    \item A potential fields-based planner that avoids local minima and reaches long-horizon goals, thanks to a geodesic attractor trained with a manifold learning objective;
    % \item An end-to-end approach that embeds the topology of the environment in a geometrically-consistent manner via self-supervised learning.
    \item An interpretable system that recovers the topology of the environment in its latent space, even in the absence of any pose information.
\end{itemize}

We show that O4A achieves state-of-the-art indoor navigation in 8 simulated environments. We further provide a real-world evaluation using the Jackal UGV platform.

\section{RELATED WORK}

The problem of robot navigation has been extensively studied \cite{thrun2005robotics}. In recent years, there has been a growing trend towards image-based learning approaches or experiential learning \cite{Levine2023survey}. These approaches have the advantage of being able to reason about the geometry of an environment, as well as the semantic aspects of traversability, such as tall grass in a field \cite{shah2021rapid}. In contrast, classical approaches that rely purely on the geometry of the environment through metric representations struggle with such semantic aspects, leading to inefficient navigation plans \cite{wang2021survey}. 

Reinforcement Learning (RL) agents are a prominent example of experiential learning and have been widely employed for image-based embodied navigation \cite{kuan2019memory, erik2019ddppo, rao2021endtoend, kulhanek2021visual}. While these approaches provide a sound solution to the navigation problem, RL policies are known to be data-hungry, with some \textit{offline} RL methods requiring up to 30 hours of training data \cite{shah2022revind}. Moreover, one of the key challenges of RL is their limited navigation capabilities over long horizons \cite{eysanbach2019sorb, dulan2021challenges}.

To overcome the long horizon limitation, semi-parametric approaches use a topological memory encoded as a graph~\cite{Kuipers1988classic}. Examples in the literature \cite{chaplot2020NTS, bansal2020combining, savinov2018sptm, wiyatno2022lifelong, shah2021ving, eysanbach2019sorb, emmons2020sparse, meng2020scaling} use this topological memory as a \textit{global planner} to obtain navigation waypoints that guide the agent towards a goal and a \textit{local policy} to navigate between these waypoints. However, these methods have two significant limitations. First, spurious connections in the graph can adversely affect the planner's ability to derive a feasible plan (e.g., they allow the planner to warp over the map). Secondly, these methods do not scale well in terms of memory as the number of vertices and edges in the graph increases with the size of the environment.

These limitations are partially addressed by hand-crafted heuristics. Examples of these include connectivity thresholds to prune the number of spurious edges \cite{savinov2018sptm, eysanbach2019sorb}, node sparsification strategies \cite{bansal2020combining, emmons2020sparse} and lifelong updates to the graph \cite{wiyatno2022lifelong}. However, these methods end up adding a non-negligible number of parameters that require tuning and are often environment dependant. In contrast, we do not rely on such heuristics and use the raw connectivity derived from an inverse kinematics head trained on top of a backbone, which is itself trained using temporal contrastive learning (i.e., explicitly maintain consecutive RGB observations close in latent space while also pushing away non-consecutive ones \cite{yang2020plan2vec}).

% In contrast, we found our method avoids the need for such heuristics by simply training a connectivity classifier on top of a backbone trained to do contrastive learning, i.e., explicitly maintain neighboring RGB observations close in a latent space while also pushing away non-neighboring observations \cite{yang2020plan2vec}. 

Of particular interest to this work are Semi-Parametric Topological Memory (SPTM) \cite{savinov2018sptm} and Visual Navigation with Goals (ViNG) \cite{shah2021ving}. SPTM uses a classifier to determine the connectivity between two temporally close images and creates an unweighted graph based on this. This graph does not accurately reflect the distance between two close or distant samples as they are not weighted, exacerbating the spurious edges issue. Similarly, ViNG uses a classifier to estimate the temporal number of steps between samples and weight the edges of the graph with these estimates. Planning is performed over the graph and a relative pose predictor is paired with a PD controller to navigate waypoints. Both methods employ pruning and sparsification strategies. Our method does not prune the graph and directly estimates the action between two samples, allowing us to predict connectivity and local controls simultaneously. 

\begin{figure*}[!ht]
\medskip
  \centering
  \includegraphics[width=\textwidth,keepaspectratio]{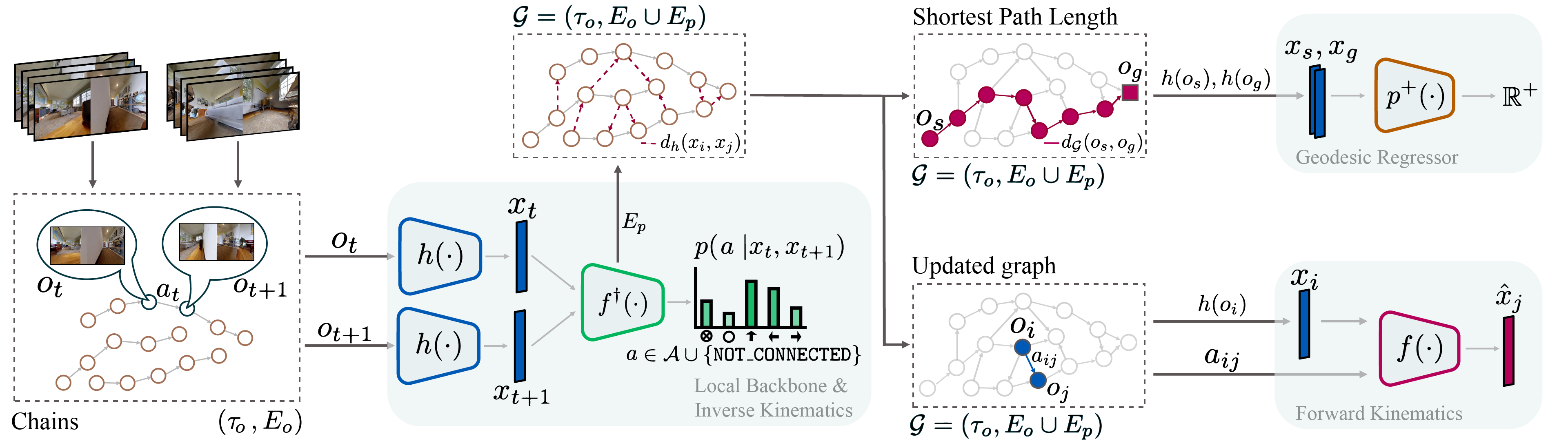}
  \caption{O4A consists of 4 learnable modules for image-goal navigation. Learning is entirely achieved using previously collected RGB observation trajectories $\tau_{\obs} =\{\obs_t\}_{t=1}^T$ and corresponding actions $\actiontraj = \{a_t\}_{t=1}^T$, without pose. The \textbf{local backbone} $\local$ (left) takes as input RGB images  to produce low-dimensional embeddings $\code \in \latentspace$. The \textbf{inverse kinematics head} $\conn$ (center) uses pairs of embeddings to predict the action required to traverse from one embedding to the other (order matters), or the inability to do so through the $\mathtt{NOT\_CONNECTED}$ output. $\local$ and $\conn$ are then used to construct a directed graph $\graph$, where nodes represent images and edges represent traversability. The \textbf{forward kinematics head} (bottom right) $\fd$ is trained using edges from $\graph$ to predict the next embedding $\code_j$ given the current one $\code_i$ and an action $a_{ij} \in \actions$. The geometry of the graph $\graph$ is embedded in a neural network using a \textbf{geodesic regressor} $\georeg$ (top right), which predicts the shortest path length for any pair of embeddings. Once all the modules are trained, $\graph$ can be discarded, and $\georeg$ be used as part of a potential function, as illustrated in Figure \ref{fig:potentials} and detailed in Equation \ref{eq:nav_cost}.}
  
  \label{fig:diagram}
\end{figure*}

Additionally, both SPTM and ViNG create a graph using \textit{all} training images and produce navigation plans by querying the graph using Dijkstra's algorithm \cite{dijkstra1959note}. In contrast, we use image embeddings to weigh a training graph and learn to predict shortest path lengths over it with a geodesic regressor. This allows to address the memory limitation of graph-based methods and also enables the estimation of shortest path distances between previously unseen images.  

There has also been research on alternative memory representations for embodied navigation, including the use of semantic \cite{kim2022topological, georgakis2022learningsemantic, Ramakrishnan2022PONIPF}, spatial \cite{gupta2017cognitive, henriques2018mapnet, chaplot2020activeslam} or embedding representations \cite{kuan2019memory}. 
% While some methods have already embedded the memory representation in the weights of a neural network \cite{kuan2019memory}, 
To the best of our knowledge, none of these representations exploit a training graph to obtain a geodesic regressor for navigation. 
Our approach is inspired by Plan2Vec \cite{yang2020plan2vec} which is a pure planning method and uses its geodesic regressor (referred to as global metric) as a heuristic for graph search. O4A instead tackles planning and navigation simultaneously and uses the geodesic regressor in a potential function \cite{choset2005principles}.
% This approach demonstrates that the geometry of the space can be estimated solely from a graph weighted with locally consistent estimates.

Finally, several existing methods rely on access to ground truth pose information in order to learn temporal estimates or relative pose between samples \cite{bansal2020combining, wiyatno2022lifelong, shah2021ving, meng2020scaling}. Our method can be trained in a self-supervised manner using only raw RGB image samples and actions, making it independent of pose information.

\section{METHOD}

\label{sec:method}
\subsection{Problem Definition}
We consider a robot with a discrete action space $\actions = \{\mathtt{STOP}, \mathtt{FORWARD}, \mathtt{ROTATE\_ RIGHT}, \mathtt{ROTATE\_ LEFT}\}$ for an image-goal navigation task \cite{anderson2018spl}. Using our knowledge of the robot's geometry and an appropriate exteroceptive onboard sensor (e.g., a front laser scanner), we assume that the set of collision-free actions $\freeactions$ can be estimated.

When prompted with a goal image $\obs_g$, the agent should navigate to the goal location in a partially observable setting using only RGB observations $\obs_t$ and the $\freeactions$ estimates. The agent further needs to identify when the goal has been reached by \textit{autonomously} calling $\mathtt{STOP}$ near the goal \cite{anderson2018spl}.

\subsection{Data}
\label{subsec:data}
% \paragraph{Data}
We aim to achieve image-goal navigation using learned modules parameterized by deep neural networks. For any given environment, we assume  that some previously collected observation trajectory $\datatraj = \{\obs_t\}_{t=1}^T$ and corresponding actions $\actiontraj = \{a_t\}_{t=1}^T$ are available. We consider a single trajectory from a single environment for notational conciseness, but in practice use multiple data trajectories (Fig. \ref{fig:diagram}) from different environments. We do not require an expert data collection policy and the dataset could be the product of teleoperation, self-exploration or random walks, as long as it sufficiently covers the free space of the environment. Of note, we tackle navigation in an unsupervised setting and do not assume access to pose estimates for each image observation, which greatly simplifies data collection. Moreover, we do not collect any depth measurement, and only rely on a front laser scanner at runtime for simple collision checking.

\subsection{System}
\label{subsec:system} 

\paragraph{Overview}
We illustrate and present an overview of our system in Figure~\ref{fig:diagram}. We first rely on self-supervised learning to learn an RGB backbone paired with a connectivity head to infer a graph over all images in $\datatraj$. The graph will then be used to derive training objectives for a forward kinematics module and a geodesic regressor. We finally show how to navigate the trained system in Subsection \ref{subsec:navigation}.
 
\paragraph{Local Backbone}
The local backbone learns a mapping from raw images to low-dimensional embeddings $\local: \rgbspace \to \latentspace $. For simplicity, we will denote extracted features as $\code = \local (\obs)$. The function $\local$ will serve a    dual purpose: 1) to extract low-dimensional features in $\latentspace = \mathbb{R}^n$ that will be used as input for other modules, and 2) to learn a \textbf{local metric} defined as
 
 \begin{equation}
 \label{eq:local_metric}
   \localmetric(\code_t, \code_s) =     \norm{\code_t - \code_s}  
 \end{equation}
 
between pairs of observations. Given the lack of pose information in the training data, $h$ is trained via self-supervised learning. We use a variant of a contrastive loss function often used to train Siamese architectures \cite{florence2018dense}

\begin{equation}
\label{eq:local_loss}
\begin{split}
\mathcal{L}_{\local}(\code_t, \;\code_{t+1},\; \mathcal{N}) &= (\posmargin - \localmetric(\code_t , \code_{t+1}))^2  \\
&+ \frac{1}{|\mathcal{N}|}\sum_{\code^{-} \in \mathcal{N}}\max(0, \negmargin - \localmetric(\code_t, \code^{-}))^2,
\end{split}
\end{equation}

\noindent
where $\posmargin,\; \negmargin \in \mathbb{R}^+, \; \posmargin < \negmargin$ are positive and negative margins, respectively, and $\mathcal{N}$ is a set of random data from $\datatraj$, which are used as so-called negatives. Equation~\ref{eq:local_loss} is an instance of temporal contrastive learning: consecutive observations (positive pairs), which we know to be indeed close in terms of pose, are encouraged to be at a distance of exactly $\posmargin$ in $\latentspace$. Negatives are pushed to be at least at a distance of $\negmargin$, reflecting the fact that they should not share the same neighborhood even if the exact distance between them is unknown at this stage. This latest observation motivates the term "local metric" \cite{yang2020plan2vec}, since the actual distance $d_{\local}$ is only informative when applied to positive pairs that are close in latent space. It should be stressed that $d_{\local}$ cannot predict how far negative pairs are apart in general, as it tends to saturate around $\negmargin$ as discussed in \cite{yang2020plan2vec}. 
%For the remainder of this work, we will make the assumption that positive samples are at most one step ahead, i.e., one-step neighbors.

 \paragraph{Inverse Kinematics Head}
The component $\conn: \latentspace \times \latentspace \to \actions \cup \{\mathtt{NOT\_CONNECTED}\} $ predicts the action needed to travel between two embeddings, or returns the $\mathtt{NOT\_CONNECTED}$ token when the transition is deemed not feasible in a single action. $\conn$ therefore acts as both a loop closing module and an inverse kinematics predictor. It is trained using the standard cross-entropy loss on the actions observed in $\actiontraj$. We use the same negatives $\mathcal{N}$ from Equation~\ref{eq:local_loss} to train the $\mathtt{NOT\_CONNECTED}$ class.  
 
 Even if most negatives in $\mathcal{N}$ are true negatives (in the sense that the observations are not connectable with one action step), both $\local$ and $\conn$ can be exposed to occasional false negatives during training. For example, if the same location is visited twice, the induced observations may not be temporally consecutive and can then appear in $\mathcal{N}$. These false negatives in fact correspond to the loop closures that should be discovered by the trained system in the data. In practice, it turns out that false negatives do not prevent $\conn$ from learning a decent connectivity (Fig. \ref{fig:globals_graphs}). 
 
 %This property allows self-supervised methods to learn useful data representations in the absence of a clean supervisory signal (for navigation, the relative pose between images. 
 % It is important to note that false negatives have the potential to bias the loss \cite{chuang2020hardnegatives}. To mitigate this issue, hard-negative mining strategies can be employed, although we leave this for future work.
 
 % \sm{add note on hard negative mining}
 %This property of self-supervised methods compensates for the lack of a clean supervisory signal (for our navigation problem, the known relative pose between two observations).

\paragraph{Graph Construction}
Equipped with $\local$ and $\conn$, we can now build a \textbf{directed graph} $\graph$ whose edges are weighted using $\localmetric$ (Equation~\ref{eq:local_metric}). We first treat the collected data as a chain graph with observed edges $E_o = \{(\obs_t, \obs_{t+1}): \obs_t, \obs_{t+1} \in  \datatraj\}$ and then run pairwise computations to obtain new loop closure edges $E_p = \{(\obs_t, \obs_{s}): \obs_t, \obs_{s} \in  \datatraj, \; \conn(\code_t, \code_s) \in \actions\}$. The final graph is $\graph = (\datatraj, E_o \cup E_p)$. No additional post-processing of the graph is required, contrary to existing methods \cite{savinov2018sptm, wiyatno2022lifelong, shah2021ving, emmons2020sparse} which can require tuning numerous hyperparameters to curate nodes and edges.

\paragraph{Forward Kinematics Head}
The forward kinematics head is denoted by $\fd: \latentspace \times \actions \to \latentspace$ and trained using edges/transitions from $\graph$. For any edge $(\obs_t, \obs_s)$ in $\graph$ during training, the module is trained with the mean squared error loss to approximate the function $(\code_t, \conn(\code_t, \code_s)) \mapsto \code_{s}$, using the inverse kinematics head $\conn$ to provide an input action even if none was observed. $\fd$ will therefore benefit from additional transitions in $E_p$ that were not initially observed in $E_o$. The above is an instance of semi-supervised learning called co-training \cite{blum1998cotraining}, in which the functions $\local$ and $\conn$ are used to label unseen transitions in the training set, thus enhancing the supervisory signal that is employed to train $\fd$.

\paragraph{Geodesic Regressor}
The final component and core planning module $\georeg: \latentspace \times \latentspace \to \mathbb{R}^{+}$ learns to predict the shortest path lengths on $\graph$. We denote these distances as $d_{\graph}(\obs_t, \obs_g)$ and compute them with Dijkstra's algorithm. $d_{G}$ is defined over observation pairs from the discrete vertex set of $\graph$. We aim to extend it over the continuous latent space $\mathcal{X}$ to predict shortest path lengths for any pair of images at runtime. The training loss of the geodesic regressor is

\begin{equation}
\label{eq:geo_reg_loss}
\mathcal{L}_{\georeg}(\obs_s, \obs_t) = (\georeg(\code_s, \code_t) - d_{\graph}(\obs_s, \obs_t))^2.
\end{equation}

%\ms{\textbf{ADD MORE STUFF HERE - INTUITION AND HIGHLIGHT WE DO NOT POST-PROCESS GRAPH}}
Interpreting observations as samples from a manifold embedded in the high-dimensional RGB space, the backbone $h$ learns an embedding with locally Euclidean neighborhoods ($d_h$), which are chained together by the graph search to compute the geodesic (intrinsic) distance over the entire manifold. Equation \ref{eq:geo_reg_loss} in fact corresponds to a manifold learning objective \cite{yang2020plan2vec, joshua2000isomap}, and we will show the O4A training results in interpretable visualizations of the environment in Figure \ref{fig:globals_graphs}.

Once all the components have been trained, $\graph$ can be discarded and is not required for deploying the system. Indeed, both $\fd$ and $\georeg$ will provide all the required information for image-goal navigation, as we will detail in Subsection \ref{subsec:navigation}. In fact, the geodesic regressor $\georeg$ can be interpreted as encoding the geometry of $\graph$, thereby trading a potentially high number of nodes and edges for a fixed number of learnable parameters.

\paragraph{Multiple Environment Setting} 

When $k$ environments are considered, we train both $\local$ and $\conn$ on the entire data. To provide a more challenging task for the model, we sample negatives $\mathcal{N}$ from either the same environment or a different one.
% \sm{Specifically, we restrict the negatives $\mathcal{N}$ to come from the same environment as $\obs_t$ to provide a more challenging task for the model.} 
$\local$ and $\conn$ can then be used to close loops and compute a set of graphs $\{\graph_i\}_{i=1}^k$, one per environment. The forward kinematics $\fd$ are then trained using transitions from all the graphs. Finally, each $\graph_i$ is used to train a geodesic regressor $\georeg_{i}$. In summary, $\local$, $\conn$ and $\fd$ are shared across environments while $\georeg_i$ is environment-specific. 

%    \sm{Given a new environment and some exploratory data from it, $\local$, $\conn$ and $\fd$ can be reused as is. The system is "finetuned" to the new geometry by computing a new graph and training a new geodesic regressor $\georeg$.}

\subsection{Navigation}
\label{subsec:navigation}
% \sm{Paragraph on Artifical Potential Fields. Comparison with navigation potential function in \cite{choset2005principles}. Discuss the geodesic aspect, which circumvents the local minima problem of potential fields. The collision checking and $\freeactions$ act as a boundary on the free configuration space.}

In this section, we discuss how to deploy O4A for navigation. Our approach is strongly inspired by Artificial Potential Field (APF) methods \cite{choset2005principles}, which plan motions over the agent configuration space by defining A) an attractive potential around the goal, and B) repulsive potentials around obstacles, allowing the agent to minimize the total potential function via gradient descent to reach the goal while avoiding obstacles.

\begin{figure}
\medskip
  \centering
  \includegraphics[width=\columnwidth,keepaspectratio]{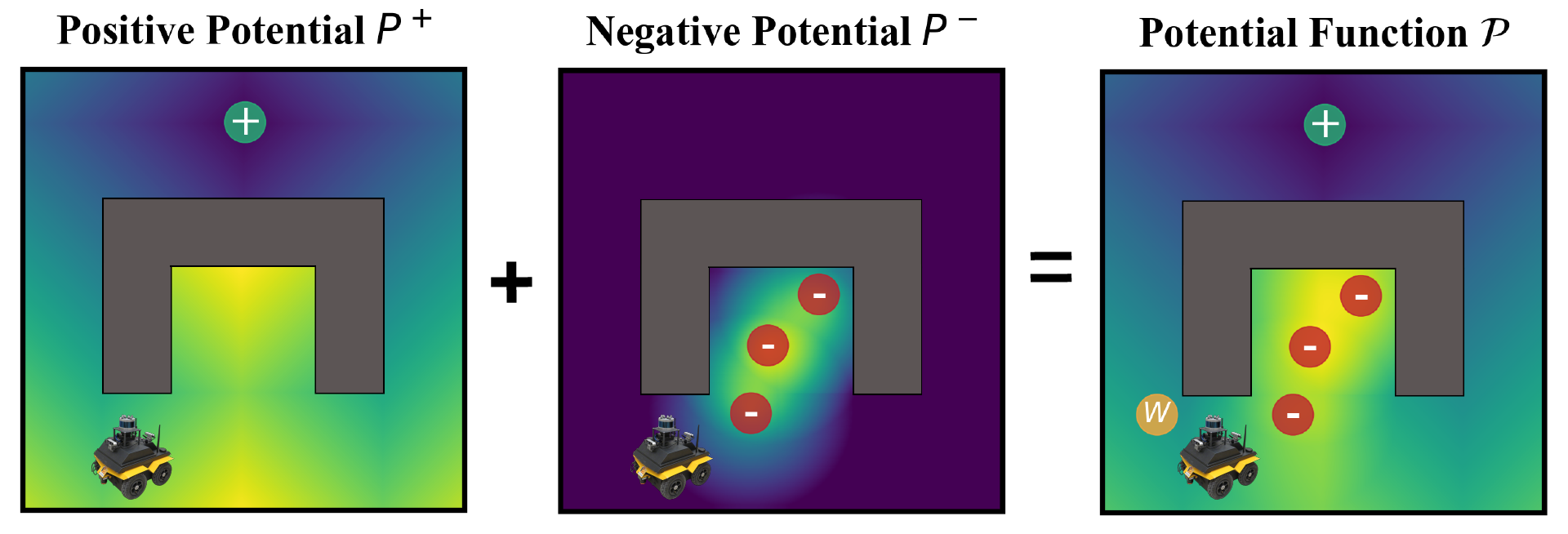}
  %\caption{A demonstration of our potential function $\mathcal{P}$ (Equation \ref{eq:nav_cost}) in a toy maze environment. In this context, areas with darker colors indicate lower potentials. From left to right, the positive potential function generates a continuous gradient whose global minima is at $G$. This potential can be utilized to determine the next waypoint $W$ from the current position $A$. The negative potential produced by previously visited states, i.e., $\{B, C, D\}$, encourages the agent to move away from past locations and prevents it from getting stuck in local minima. The combination of positive and negative potentials produces a minimization objective that drives the agent towards the goal.}
  \caption{An illustration of our potential function $\mathcal{P}$ (Equation~\ref{eq:nav_cost}). Darker colors indicate lower potential. (Left) \textbf{Geodesic attractor}, which reflects the geodesic distance to the \textcolor{ForestGreen}{green goal marker}. (Center) \textbf{Repulsors} around previously visited states represented as \textcolor{red}{red markers}. (Right) \textbf{Total potential function} $\mathcal{P}$, which is minimized by the agent by picking the action that leads to a potential-minimizing waypoint W. While we illustrate the potential function on the map, it is in fact defined directly over image embeddings.
  %The geodesic a In this context, areas with darker colors indicate lower potentials. From left to right, the positive potential function generates a continuous gradient whose global minima is at $G$. This potential can be utilized to determine the next waypoint $W$ from the current position $A$. The negative potential produced by previously visited states, i.e., $\{B, C, D\}$, encourages the agent to move away from past locations and prevents it from getting stuck in local minima. The combination of positive and negative potentials produces a minimization objective that drives the agent towards the goal.
  }
  \label{fig:potentials}
\end{figure}

As with APF, O4A will navigate by minimizing an attractor located at at the goal. Since actual agent and goal states are unobserved, the potential computations occur over the latent space $\latentspace$, i.e. the embeddings of the agent and goal RGB observations. As the attractor, we use the geodesic regressor $\georeg$ which estimates the geodesic distance to the goal. Critically, this attractor factors in the environment geometry and can, for example, drive an agent out of a dead end to reach a goal that is close in terms of Euclidean distance, but far geodesically (c.f. Fig. \ref{fig:potentials}). This property is somewhat reminiscent of navigation functions in APF literature \cite{choset2005principles}, a special class of potential functions with a unique global minimizer at the goal, among other properties.

\input{figures/navigation.tex}

In practice, we found that minimizing $\georeg$ alone did not suffice to successfully navigate. The agent would often end up thrashing between two poses due to a local minimum in the attractor landscape, which can occur due to learning errors and the discrete action space. We therefore found it useful to define a latent repulsor function, which is only active in a certain radius $m_r \in \mathbb{R}^+$ :
\begin{equation}
    p^-(\code_t, \code_s) = \max(0, m_r - \localmetric(\code_t, \code_s)).
\end{equation}

We use $p^-$ to drive the agent away from previously visited images, the embeddings of which are saved in a buffer $\mathcal{B}$. By combining repulsors around embeddings in $\mathcal{B}$ and the geodesic attractor $\georeg$, we obtain a total potential function of

\begin{equation}
\label{eq:nav_cost}
    \mathcal{P}(\code_t, \code_g, \mathcal{B}) = \georeg(\code_t, \code_g) + \sum_{\code \in \mathcal{B}}p^-(\code_t, \code)
\end{equation}

\noindent
where $\code_t$ and $\code_g$ represent the embeddings of the current and goal RGB images, respectively.

 The detailed navigation procedure is presented in Algorithm \ref{algo:navigation}. During navigation, our agent greedily minimizes $\mathcal{P}$ by finding the best candidate action using forward kinematics over the set $\freeactions$ estimated by a collision detection function $\gamma$. This stands in contrast to APFs, since we blacklist collision-inducing actions instead of explicitly modeling repulsors around obstacles. In practice, since the agent rotates in place, we suppose that only the $\mathtt{FORWARD}$ action can cause collisions, which greatly simplifies the collision detection $\gamma$: we simply define a scan collision box in front of the robot based on its geometry. 
 % Our approach is similar to the one proposed by \cite{mishkin2019benchmark}, with the difference that we do not estimate a two-dimensional obstacle map from the laser scan. Instead, we use this measurement to identify immediate forward collisions.

It should also be noted that the $\mathtt{STOP}$ action is never included in $\freeactions$. Instead, we found that thresholding the local metric $d_{\local}$ was a more reliable way of calling $\mathtt{STOP}$ in the vicinity of the goal.

\section{EXPERIMENTS}
\label{sec:experiment}
We assess our approach in both simulated and real-world environments. The agent is a differential drive robot with two RGB cameras, one facing forward and the other facing backward, each with a field of view of $90^\circ$. Each image has a resolution of $96 \times 96$ pixels. Consistent with \cite{hahn2021norl}, the robot moves $\mathtt{FORWARD}$ by 0.25m and $\mathtt{ROTATE}$ by $15^\circ$. 

\subsection{Implementation Details}
\label{subsec:implementation}

\input{figures/main_table}

All our models are trained using a batch size of 512. The local backbone uses a ResNet 18 encoder \cite{he2016resnet}, followed by 5 1D convolutions to fuse the embeddings of the front and rear facing cameras. The inverse kinematics head is composed of 4 linear layers. Both the local backbone and inverse kinematics heads are jointly trained for 410,000 gradient steps, using $\posmargin=1$ and $\negmargin=10$. The forward kinematics model has 4 linear layers and is trained for 330,000 steps. The geodesic regressor has 6 linear layers, and the same architecture is used for all environments, with training also carried out for 330,000 steps for each regressor. All models are trained using image augmentations and linear layers are followed by ReLU non-linearities. The models are trained using the Adam optimizer \cite{kingma2014adam} with learning rate $5e^{-4}$.  Navigation is performed with $|\mathcal{B}| = 500$, $m_r = 2.5$ and $\mathtt{thresh} = 3.5$ throughout environments and experiments.

\subsection{Simulation}
\label{subsec:simulation_results}

\paragraph{Simulator \& Data}

We perform our experiments using the Habitat simulator~\cite{manolis2019habitat} with the Gibson dataset~\cite{xia2018gibson}. We use the Gibson split defined in \cite{manolis2019habitat} and use eight environments from it: Hambleton (67$m^2$), Annawan (75$m^2$), Nicut (90$m^2$), Dunmor (90$m^2$), Cantwell (107$m^2$), Sodaville (114$m^2$), Aloha (114$m^2$) and Eastville (121$m^2$). A total of 240,000 data points are collected by navigating sequences of random nearby waypoints, leading to globally suboptimal trajectories between the initial and final positions. We split the data into a training set (70\%) and a validation set (30\%).

% Local is actually 63K

\paragraph{Baselines}

To assess the navigation performance of our method, we compare against the following baselines:

\begin{itemize}
    \item \textbf{Random Agent.} An agent sampling actions uniformly from $\actions\setminus\{\mathtt{STOP}\}$, subject to collision checking. The policy relies on ground truth pose for oracle stopping.
    \item \textbf{Goal Conditioned Behavioral Cloning (GC-BC)}, adapted from \cite{codevilla2018goalbehavior}. We extended the method with Hindsight Experience Replay (HER) \cite{andrychowicz2017her} to relabel new goals and enhance the training signal for the agent. This policy also relies on ground truth pose to call $\mathtt{STOP}$. We collected a distinct set of expert trajectories for this baseline.
    \item \textbf{SPTM} \cite{savinov2018sptm}. This method is extended with hard-negative mining as proposed by \cite{shah2021ving} to improve generalization performance. To call $\mathtt{STOP}$, we tune a threshold on the reachability estimator between current and goal images.
    \item \textbf{ViNG} \cite{shah2021ving}. The PD controller of the original work is replaced by an oracle Habitat controller to handle discrete actions. We tune a threshold on the timestep predictor to call $\mathtt{STOP}$.
    \item \textbf{O4A Without Latent Repulsors}. The repulsors $p^-$ are not used during navigation and the agent is driven towards the goal by greedily minimizing $p^+$.
    \item \textbf{O4A Without Geodesic Regressor}. We discard the attractor $\georeg$ and only use the latent repulsors $p^-$.
\end{itemize}

To ensure a fair comparison, all the baselines have the same capacity and collision checking strategy as O4A. We substitute the neural architectures used in the baselines with our local backbone $\local$ and inverse kinematics head $\conn$. Baselines are trained for 415,000 gradient steps and tested with various hyperparameters, the best of which were used for benchmarking.

\paragraph{Evaluation}

To assess the navigation performance, we rank the difficulty of each trajectory as proposed by \cite{chaplot2020NTS}. Trajectories are categorized into easy (1.5 $-$ 3m), medium (3 $-$ 5m), and hard (5 $-$ 10m) based on their geodesic distance to the goal. We add an extra category labeled as "very hard" ($>$10m) to evaluate the agent's ability to plan for long-horizon goals. Each difficulty level is assessed over 1,000 trajectories using a maximum episode length of 500. To ensure a comprehensive assessment, we sample distinct starting positions and goals for each environment and difficulty, resulting in a total of 4,000 trajectories per environment. Starting and goal positions used for evaluation are different from those used during data collection.

We assess navigation performance by measuring Success Rate (SR) and Success weighted by Path-Length (SPL) \cite{anderson2018spl}. A navigation trial is considered successful if the agent comes to a $\mathtt{STOP}$ within a maximum distance of 1m from the goal. We also use a \textit{Soft} Success Rate (SSR), where a trial is successful if the agent is less than 1m away from the goal at any point during navigation. 
% To assess collision checking, we measure the average number of collisions per trajectory, which is given by $\frac{1}{T}\sum_i^{T} c_i$, where $c_i$ counts the number of collisions per trajectory $i$. 
Lastly, we monitor the ratio of Collision-Free Trajectories (CFT) across all difficulties. A trajectory is deemed collision-free if it does not collide during the experiment.
% by calculating $\frac{1}{T}\sum_i^{T} \mathbbm{1}(i)$, where the indicator function $\mathbbm{1}(\cdot)$ is equal to one if trajectory $i$ is collision-free.

\paragraph{Results}
\label{para:sim_results}

We present quantitative navigation results in Table \ref{table:table_main}. O4A successfully navigates to goals of all difficulty levels and outperforms all considered baselines. The tight gap between SR and SSR showcases the reliability of the $\mathtt{STOP}$ mechanism based on the local metric $d_h$. Surprisingly, the O4A SPL is stable across all difficulties, despite a decreasing success rate. This indicates that the quality of successful paths is in fact slightly better for distant goals. This observation, combined with visual evaluation of episodes, suggests that the O4A attractor provides a clearer signal for distant goals and is noisier when navigating nearby locations. The two O4A ablations confirm that all considered potentials in our potential function $\mathcal{P}$ are essential contributors to success. 

% \vspace{-3.5mm}

\begin{figure}[!htbp]
    \centering
    % \begin{subfigure}[b]{\columnwidth}
    %     \centering
    %     \includegraphics[width=0.49\columnwidth]{figures/annawan_graph.png}%
    %     \includegraphics[width=0.49\columnwidth]{figures/annawan_global.png}
    %     \caption{Annawan}
    % \end{subfigure}
    % \vskip\baselineskip
    
    \begin{subfigure}[b]{\columnwidth}
        \centering
        \includegraphics[width=0.475\columnwidth]{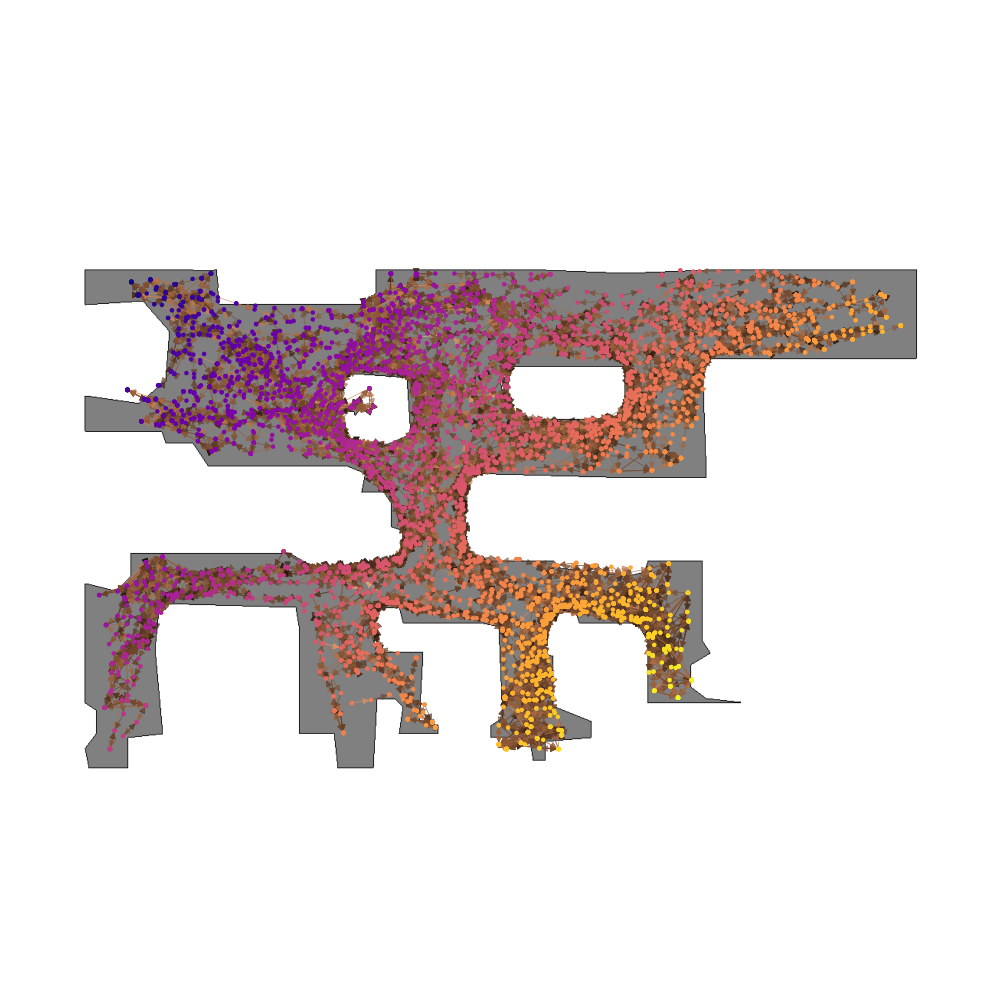}%
        \includegraphics[width=0.475\columnwidth]{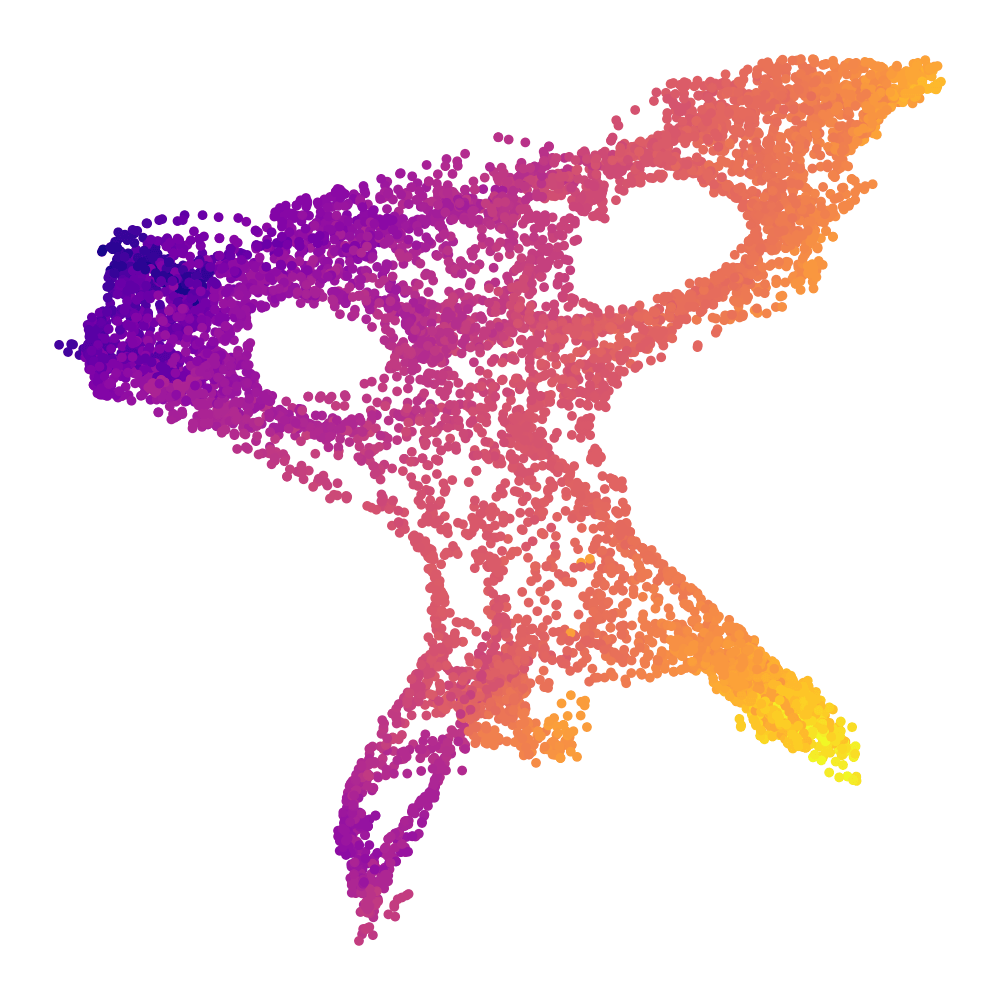}
        \caption{Hambleton}
        % \vspace{-0.25mm}
    \end{subfigure}
    % \vskip\baselineskip
    \begin{subfigure}[b]{\columnwidth}
        \centering
        \includegraphics[width=0.475\columnwidth]{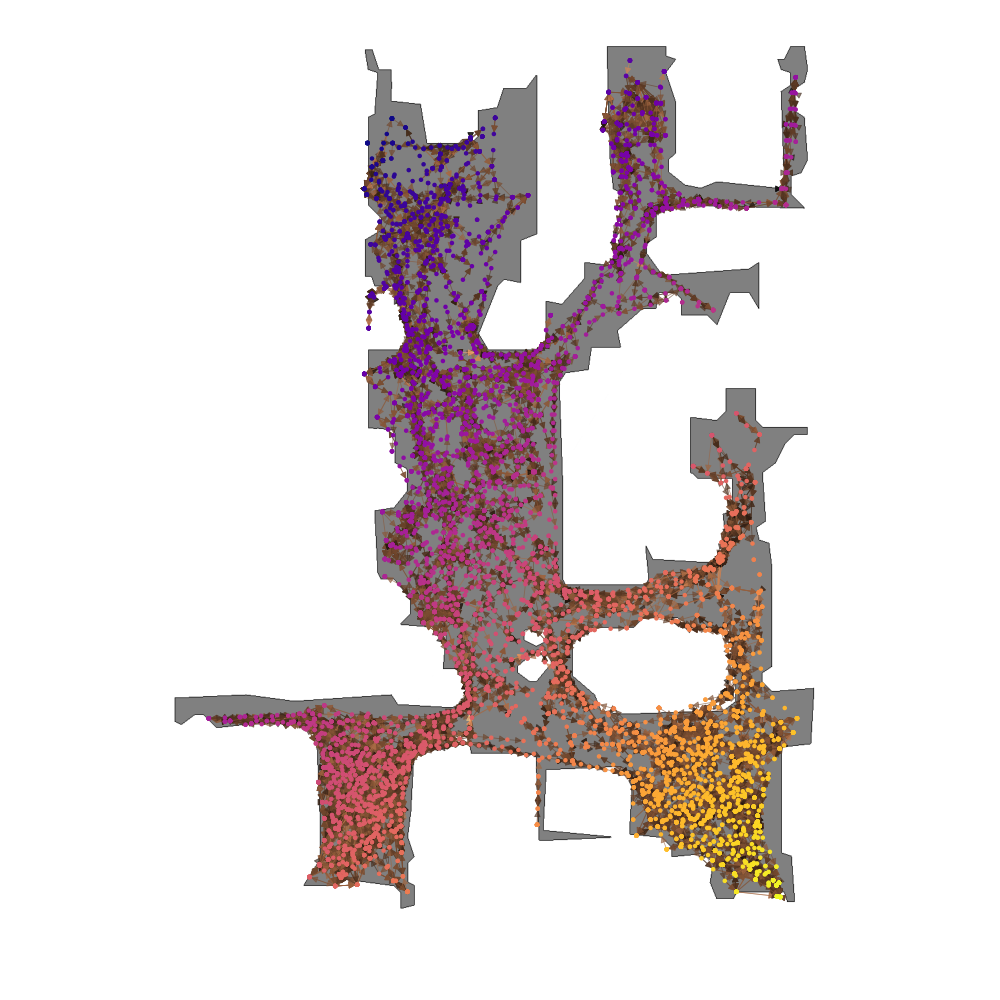}%
        \includegraphics[width=0.475\columnwidth]{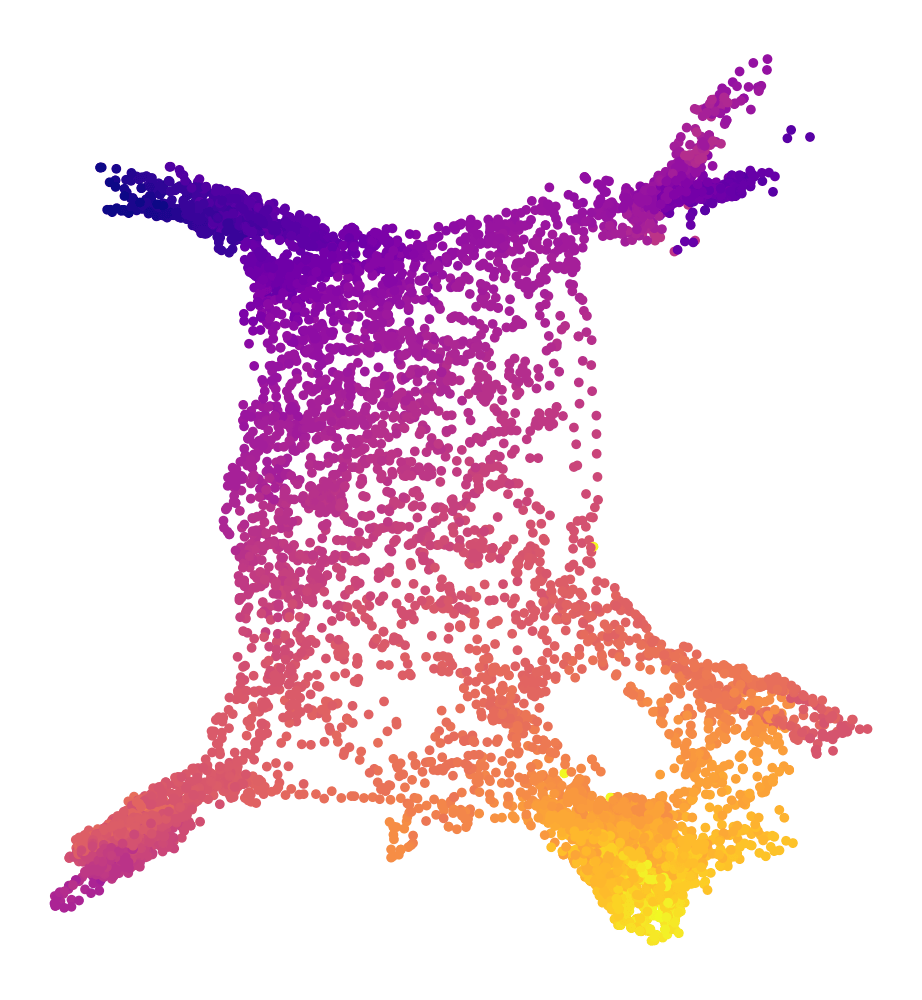}
        \caption{Eastville}
        % \vspace{-0.25mm}
    \end{subfigure}
    
    % \caption{Geometric consistency of our model with the environment. The left column shows the environment in gray along with samples plotted based on their ground truth coordinates. A graph is also shown, where the connectivity is derived from 
    % $\local$ and $\conn$. In the right column, the two first principal components of the embeddings in the last layer of $\georeg$ are shown. Our model exhibits consistency with the topology of the environment, even without explicitly using pose information.}
    \caption{(Left column) O4A graph connectivity over 2 Gibson environments. Points correspond to the location of RGB observations and are colored by the sum of their $x$ and $y$ coordinates. The graphs are free of egregious spurious edges, which allows to train effective geodesic regressors before discarding them. (Right column) 2 principal components of the last layer in the geodesic regressor $\georeg$ with the same coloring scheme. The unsupervised latent geometry is consistent with the environment geometry, and some topological features (e.g., the obstacle "holes") are evident in the latent space, even if the training of O4A never used pose information.}
    \label{fig:globals_graphs}
\end{figure}

\vspace{-2.5mm}

While the SPTM and ViNG performance are below those reported in the original papers, they are in line with recent comparable benchmarks in the literature \cite{wiyatno2022lifelong}. The gap can be partially explained by variations in the experimental design. Neither method considers collisions nor how to stop when the goal is reached. Having $\mathtt{STOP}$ in the action space increases the challenge, due to the agent's ability to prematurely terminate episodes because of a false positive. We also note that SPTM considered omnidirectional actions, and that the original ViNG results focused on larger-scale problems in relatively open outdoor settings. The GC-BC results demonstrate how learning long-range navigation remains challenging for end-to-end methods.

On a more qualitative note, we further explore the O4A graph connectivity used for training, and show how it learns interpretable embeddings in Figure \ref{fig:globals_graphs}.

% \begin{figure}
%   \centering
%   \includegraphics[width=0.60\columnwidth,keepaspectratio]{figures/jackal_no_bg.png}
%   \caption{We use the Jackal UGV mobile platform using the RGB channels (no depth) of two Realsense D435i cameras with a $90^\circ$ FOV. We detect forward collisions using a forward-facing Hokuyo laser scanner.  We run O4A onboard using an Intel i7-8700 CPU with 32 GB of RAM. We did not require a GPU for navigation.
%   %The vehicle is equipped with a GPS, wheel encoders and a 6-DoF IMU unit although these are not used in our experiments. For the forward and , but we do not use the depth component.  
%   }
%   \label{fig:jackal}
% \end{figure}
% \vspace{-2.5mm} % KABOOM NOPE YES
% \vspace{-20.0mm}
\subsection{Real-World Results}
\label{subsec:robot_results}

\paragraph{Robot Evaluation}

For the real-world experiments we run O4A on the Jackal UGV mobile platform using the RGB channels (no depth) of two Realsense D435i cameras with a $90^\circ$ FOV. We detect forward collisions using a forward-facing Hokuyo laser scanner.  We run O4A onboard using an Intel i7-8700 CPU with 32 GB of RAM. We did not require a GPU for navigation. The experiments are conducted in a 4.65m x 9.1m laboratory.
%For computing navigation metrics, we track the robot and goal poses using a ViCON system. This information is not available to the agent and is solely used for evaluation. 
We collected a total of 21,000 RGB image samples via teleoperation and split them as in our simulation experiments.

For evaluation, we chose 9 interesting episodes (3 starting positions with 3 goal images each) and repeated each 3 times, for a total of 27 runs. The maximum episode length was set to 300 steps, although the robot ended up exceeding 150 steps on only 2 occasions. In addition to SR and SSR (Subsection \ref{subsec:simulation_results}), we evaluate the final distance to goal (DTG), the number of $\mathtt{FORWARD}$ steps, and the number of $\mathtt{ROTATION}$ steps. For context, we also teleoperated the robot over the same episodes to provide an estimate of human performance. All the models used for the real-world experiments are finetuned for 30,000 gradient steps starting from the best checkpoint obtained in the simulation experiments. We run navigation with $|\mathcal{B}| = 300$, $m_r = 2.5$ and $\mathtt{thresh} = 5$.

\paragraph{Results}

The Jackal navigation results are presented in Table \ref{table:table_jackal}. O4A solves most episodes and achieves an average DTG of under 1m, even if most goals were not visible from the starting location and located up to 9 meters away. The maximal measured DTG was 1.74m. Interestingly, the number of O4A calls to the $\mathtt{FORWARD}$ action is comparable to human performance, meaning the O4A paths over the plane are competitive. Moreover, the robot collided only one time over all episodes using our collision checking strategy.  Two episodes are shown in Figure~\ref{fig:robot_trajs}.

% \vspace{5.0mm}
\input{figures/jackal_table.tex}

 \section{Limitations  \& Future Work}
%We presented \textbf{One-4-All}, a deep-learning based approach for learning image-goal navigation tasks. 
%offline from sequences of RGB observations and controls, without requiring any pose or depth measurements. 
While trained O4A is graph-free, we still require learning a geodesic regressor for each environment to encode the geometry (in the same way current approaches need to build an environment-specific graph). Generalizing geodesic regression across environments is a promising area of research, since it could allow to completely skip the graph building stage in new settings. Moreover, the real-world experiments reveal that O4A has difficulty minimizing the number of rotation actions and some amount of trashing persists. We believe that this may be caused by the $15^\circ$ discrete rotation actions: if the robot ideally needs to turn by $7.5^\circ$, it may instead oscillate between $\mathtt{LEFT}$ and $\mathtt{RIGHT}$ due to the fact actions are greedily taken each step instead of explicitly following a long-term plan. Further tuning of the negative potentials or an implementation with a continuous action space should address this problem. Finally, as with a number of existing navigation and SLAM systems, O4A does not account for dynamic or semi-static objects.

%Finally, the strict reliance on RGB data of our approach greatly simplifies data collection, meaning it could be potentially scaled to large-scale datasets \cite{shah2022lmnav} combining both indoor and outdoor data. 
% A conclusion section is not required. Although a conclusion may review the main points of the paper, do not replicate the abstract as the conclusion. A conclusion might elaborate on the importance of the work or suggest applications and extensions. 

\addtolength{\textheight}{-12cm}   % This command serves to balance the column lengths

%%%%%%%%%%%%%%%%%%%%%%%%%%%%%%%%%%%%%%%%%%%%%%%%%%%%%%%%%%%%%%%%%%%%%%%%%%%%%%%%

%%%%%%%%%%%%%%%%%%%%%%%%%%%%%%%%%%%%%%%%%%%%%%%%%%%%%%%%%%%%%%%%%%%%%%%%%%%%%%%%


% \input{figures/hyperparams_table.tex}

% Appendixes should appear before % the acknowledgment.

\section*{ACKNOWLEDGMENTS}
The authors would like to thank Kaustubh Mani, Simon Chamorro, Ali Harakeh and Steven Parkison for their help and comments. This research was partially funded by an IVADO MSc. Scholarship and an FRQNT B1X Scholarship [S.M.]. The work was also supported by CIFAR under the Canada-CIFAR AI Chair program, as well as the NSERC Discovery Grant (RGPIN-2018-04653). Moreover, this research was enabled in part by support provided by Calcul Québec and the Digital Research Alliance of Canada.

\bibliographystyle{IEEEtran}
\bibliography{ref-ab, ref}

\end{document}

%% file: figures/navigation.tex
\begin{algorithm}[h]
\caption{Navigation}
\label{algo:navigation}

 \hspace*{\algorithmicindent} \textbf{Input}: env, goal image $\obs_g \in \rgbspace$, $\mathtt{thresh} \in \mathbb{R}^+$, backbone $\local$, forward kinematics $\fd$, collision detection function $\gamma$, potential function $\mathcal{P}$ 
\begin{algorithmic}
\setstretch{1.2}

%Initialization
\State $\obs$, $\;\mathtt{scan} \gets \mathtt{env.initialize()}$
\Comment{RGB and scan}
\State $\code \gets \local(\obs)$
\State $\code_g \gets \local(\obs_g)$
\State $\mathcal{B} \gets \{\code\}$

\While {$d_h(\code, \code_g) > \mathtt{thresh} $}
\State $\freeactions \gets \gamma(\mathtt{scan})$
\State $a^* \gets \argmin\limits_{a \in \freeactions} \; \mathcal{P}(\fd(\code, a), \code_g, \mathcal{B})$
\Comment{Eq. \ref{eq:nav_cost}}
\State $\obs$, $\;\mathtt{scan} \gets \mathtt{env.step(a^*)}$
\Comment{RGB and scan}
\State $\code \gets \local(\obs)$
\State $\mathcal{B} \gets \mathcal{B} \cup \{\code\}$
\Comment{Update visited states}

\EndWhile
\State $\mathtt{env.step(STOP)}$
\end{algorithmic}
\end{algorithm}

%% file: figures/main_table.tex
\begin{table*}
\medskip
\centering
\caption{Average navigation performance over 8 simulated Gibson environments for One-4-all (O4A) and relevant baselines. We further study two additional variants of O4A by ablating terms in the potential function (Equation \ref{eq:nav_cost}). We also denote which methods rely on a graph ($\graph$) for navigation and oracle stopping (other methods need to call $\mathtt{STOP}$ autonomously). We find that O4A substantially outperforms baselines, achieving a higher Success Rate (SR), Soft Success Rate (SSR), Success Weighted by Path Length (SPL), and a competitive ratio of Collision-Free Trajectories (CFT). }
\setlength{\tabcolsep}{4.5pt}
\begin{tabular}{lccccccccccccccc}
\toprule
 & \multirow{2}[3]{*}{$\mathcal{G}$} & \multirowcell{2}{Oracle \\Stop} & \multicolumn{3}{c}{Easy (1.5 - 3m)} & \multicolumn{3}{c}{Medium (3 - 5m)} & \multicolumn{3}{c}{Hard (5 - 10m)} & \multicolumn{3}{c}{Very-Hard ($>$10m)} & \multirow{2}[3]{*}{CFT$\uparrow$} \\
\cmidrule(lr){4-6} \cmidrule(lr){7-9} \cmidrule(lr){10-12} \cmidrule(lr){13-15}
& & & SR$\uparrow$ & SSR$\uparrow$ & SPL$\uparrow$ & SR$\uparrow$ & SSR$\uparrow$ & SPL$\uparrow$ & SR$\uparrow$ & SSR$\uparrow$ & SPL$\uparrow$ & SR$\uparrow$ & SSR$\uparrow$& SPL$\uparrow$ \\
\midrule
Random & \textrm{--} & $\checkmark$ &  0.49 & 0.49 & 0.32 &  0.24 & 0.24 & 0.15 &  0.07 & 0.07 & 0.04 &  0.01 & 0.01 & 0.01 & 0.94 \\
GC-BC \cite{codevilla2018goalbehavior} & \textrm{--} & $\checkmark$ & 0.42  & 0.42 & 0.25 & 0.21 & 0.21 & 0.10 & 0.07 & 0.07 & 0.03 & 0.02 & 0.02 & 0.01 & 0.94\\
SPTM \cite{savinov2018sptm} & $\checkmark$ & \textrm{--} & 0.32 & 0.51 & 0.14 & 0.17 & 0.28 & 0.06 & 0.07 & 0.12 & 0.03 & 0.02 & 0.03 & 0.01 & 0.96\\
ViNG \cite{shah2021ving} & $\checkmark$ & \textrm{--} & 0.29 & 0.64 & 0.10 & 0.19 & 0.46 & 0.07 & 0.11 & 0.28 & 0.05 & 0.06 & 0.12 & 0.02 &\textbf{0.99}\\
O4A (Ours) & \textrm{--} & \textrm{--} & \textbf{0.95} & \textbf{0.97} & \textbf{0.65} & \textbf{0.93} & \textbf{0.95} & \textbf{0.65} & \textbf{0.90} & \textbf{0.92} & \textbf{0.65} & \textbf{0.85} & \textbf{0.88} & \textbf{0.65} & 0.96\\
\midrule
O4A w/o $p^-$ & \textrm{--} & \textrm{--} & 0.11 & 0.36 & 0.11 & 0.07 & 0.22 & 0.07 & 0.03 & 0.09 & 0.03 & 0.01 & 0.02 & 0.01 & \textbf{0.99}\\
O4A w/o $\georeg$ & \textrm{--} & \textrm{--} & 0.45 & 0.70 & 0.13 & 0.27 & 0.47 & 0.09 & 0.13 & 0.23 & 0.04 & 0.04 & 0.07 & 0.02 & 0.90 \\
 
\bottomrule
% \vspace{-7.0mm}
\end{tabular}
\label{table:table_main}
\end{table*}

%% file: figures/jackal_table.tex
\begin{table}[!hptb]
\medskip
\centering
\caption{Average Jackal navigation performance for O4A and human teleoperation over 27 episodes. We report the Success Rate (SR), Soft Success Rate (SSR), final Distance to Goal (DTG), number of $\mathtt{FORWARD}$ steps and number of rotation steps ($\mathtt{ROT.}$). %O4A reliably reaches the goals and achieves an overall low DTG and number of forward steps.
}
\setlength{\tabcolsep}{5.5pt}
\begin{tabular}{lccccc}
\toprule
& SR $\uparrow$ & SSR $\uparrow$ & DTG $\downarrow$ & $\mathtt{FORWARD}$ $\downarrow$ & $\mathtt{ROT.} \downarrow$  \\
\midrule
O4A & 0.74 & 0.78 & 0.83 & 23 & 53\\
Human & \textbf{0.96} & \textbf{1.00} & \textbf{0.46} & \textbf{20} & \textbf{18}\\
 \bottomrule
\end{tabular}
\label{table:table_jackal}
\end{table}